\documentclass{article}
\usepackage{ijcai21}
\usepackage{natbib}

\usepackage{times}
\usepackage{soul}
\usepackage{url}
\usepackage{hyperref}
\hypersetup{
colorlinks=true, 
breaklinks=true,
urlcolor= cyan, 
linkcolor= magenta,
citecolor=purple, 
}
\usepackage[utf8]{inputenc}
\usepackage[small]{caption}
\usepackage{graphicx}
\usepackage{amsmath,nicefrac}
\usepackage{amsthm}
\usepackage{booktabs}
\usepackage{algorithm}
\usepackage{algorithmic}
\usepackage{amssymb}
\usepackage{todonotes}
\usepackage{amsmath,amsfonts,mathrsfs,amssymb,bm,mathtools,mathabx}
\usepackage{multirow,mdframed}
\newcommand{\rev}[1]{{\color{purple}{#1}}}
\newcommand{\rem}[1]{{\color{cyan}{[RM?]--#1}}}

\newcommand{\nando}[1]{\todo[inline,caption={},color=magenta!20!]{Nando: #1}}

\DeclareMathOperator*{\argmin}{argmin}

\title{End-to-End Constrained Optimization Learning: A Survey}

\author{
James Kotary$^1$\and
Ferdinando Fioretto$^1$\and
Pascal Van Hentenryck$^{2}$\And
Bryan Wilder$^3$
\affiliations
$^1$Syracuse University\\
$^2$Georgia Institute of Technology\\
$^3$Harvard University
\emails
\{jkotary, ffiorett\}@syr.edu,
pvh@gatech.edu,
bwilder@g.harvard.edu
}

\begin{document}

\sloppy\allowdisplaybreaks\maketitle

\begin{abstract}
This paper surveys the recent attempts at leveraging machine learning to solve constrained optimization problems. 
It focuses on surveying the work on integrating combinatorial solvers and optimization methods with machine learning architectures. 
These approaches hold the promise to develop new hybrid machine learning and optimization methods to predict fast, approximate, solutions to combinatorial problems and to enable structural logical inference.
This paper presents a conceptual review of the recent advancements in this emerging area. 
\end{abstract}

\section{Introduction}

\emph{Constrained optimization} (CO) has made a profound impact in industrial and societal applications in numerous fields, including transportation, supply chains, energy, scheduling, and the allocation of critical resources. The availability of algorithms to solve CO problems is highly dependent on their form, and they range from problems that are efficiently and reliably solvable, to problems that provably have no efficient method for their resolution. 
Of particular interest in many fields are \emph{combinatorial optimization} problems, which are characterized by discrete state spaces, and whose solutions are often combinatorial in nature, involving the selection of subsets, permutations, paths through a network, or other discrete structures to compose a set of optimal decisions. They are known for their difficulty, and are often NP-Hard.

Despite their complexity, many CO problems are solved routinely and the AI and Operational Research communities have devised a wide spectrum of techniques and algorithms to effectively leverage the problem structure and solve many hard CO problems instances within a reasonable time and with high accuracy. While this success has made possible the deployment of CO solutions in many real-world contexts, the complexity of these problems often prevent them to be adopted in contexts of repeated (e.g., involving expensive simulations, multi-year planning studies) or real-time nature, or when they depend in nontrivial ways on empirical data. 
However, in many practical cases, one is interested in solving problem instances sharing similar patterns. Therefore, machine learning (ML) methods appear to be a natural candidate to aid CO decisions and have recently gained traction in the nascent area at the intersection between CO and ML.



Current research areas in the intersection of CO and ML can be categorized into two main directions: {\emph{ML-augmented CO}} and {\emph{End-to-End CO learning}}. 
The former focuses on using ML to aid the decisions performed within an optimization algorithm used to solve CO problems. The latter involves the combination of ML and CO techniques to form integrated models which predict solutions to optimization problems. The survey subdivides this context to into two main application domains: {\bf (1)} the fast, approximate prediction of solutions to CO problems and {\bf (2)} the integration of data-driven inference with CO solvers for structure logical inference.

While there exists work surveying ML-augmented CO methods \citep{bengio2020machine,Lodi2017Jul}, the more modern end-to-end CO learning methods lack of a cohesive and critical analysis. The goal of this survey is to address this gap and provide a focused overview on the work to-date on end-to-end CO learning, provide a critical analysis, and pose a set of open questions and directions.

\section{Preliminaries: Constrained Optimization}
A constrained optimization (CO) problem poses the task of minimizing an \emph{objective function} $f: \mathbb{R}^n \to \mathbb{R}_+$ of one or more variables $\bm{z} \in \mathbb{R}^n$, subject to the condition that a set of \emph{constraints} $\mathcal{C}$ are satisfied between the variables: 
\begin{equation}
\label{eq:opt}
\mathcal{O} = \argmin_{\bm{z}} f(\bm{z}) \;\; 
  \text{subject to} \;\;
  \bm{z} \in \mathcal{C}.
\end{equation}
An assignment of values $\bm{z}$ which satisfies $\mathcal{C}$ is called a \emph{feasible solution}; if, additionally $f(\bm{z}) \leq f(\bm{w})$ for all feasible $\bm{w}$, it is called an \emph{optimal solution}.

A well-understood class of optimization problems are \emph{convex} problems, those in which the constrained set ${\cal C}$ is a convex set, and the objective $f$ is a convex function. Convex problems have the favorable properties of being efficiently solvable with strong theoretical guarantees on the existence and uniqueness of solutions \citep{boydconvex}. 

A particularly common constraint set arising in practical problems takes the form 
$ {\cal{C}}= \{ \bm{z}  \;:\; \bm{A} \bm{z} \leq \bm{b} \}$,
where $\bm{A} \in \mathbb{R}^{m \times n}$ and $\bm{b} \in \mathbb{R}^m$. 
In this case, ${\cal{C}}$ is a convex set. If the objective $f$ is an affine function, the problem is referred to as \emph{linear program} (LP). When a linearly constrained problem includes a quadratic objective rather than a linear one, the result is called a \emph{quadratic program} (QP). 
If, in addition, some subset of a problem's variables are required to take integer values, it is called \emph{mixed integer program} (MIP). While LP and QP with convex objectives belong to the class of convex problems, the introduction of integral constraints ($\bm{x} \in \mathbb{N}^n$) results in a much more difficult problem. The feasible set in MIP consists of distinct points in $\bm{x} \in \mathbb{R}^n$, not only nonconvex but also disjoint, and the resulting problem is, in general, NP-Hard. 
Finally, nonlinear programs (NLPs) are optimization problems where some of the constraints or the objective function are nonlinear. Many NLPs are nonconvex and can not be efficiently solved \citep{nocedal2006numerical}.

Of particular interest are the \emph{mixed integer linear programs} (MILPs), linear programs in which a subset of variables required to take integral values. 
This survey is primarily concerned with optimization problems involving linear constraints, linear or quadratic objective, and either continuous or integral variables, or a combination thereof.


\paragraph{Optimization Solving Methods}

A well-developed theory exists for solving convex problems. 
Methods for solving LP problems include \emph{simplex} methods \citep{dantzig1951maximization}, \emph{interior point} methods \citep{boydconvex} and \emph{Augmented Lagrangian} methods \citep{hestenes1969multiplier,powell1969method}.
Each of these methods has a variant that applies when the objective function is convex. Convex problems  \citep{boydconvex} are in the class of $\bm{P}$, and can be assumed to be reliably and efficiently solvable in most contexts. For a review on convex optimization and Lagrangian duality, the interested reader is referred to \citep{boydconvex}.

MILPs require a fundamentally different approach, as their integrality constraints put them in the class of NP-Hard problems. 
\emph{Branch and bound} is a framework combining optimization and search, representable by a search tree in which a LP \emph{relaxation} of the MILP is formed at each node by dropping integrality constraints, and efficiently solved using solution methods for linear programming. Subsequently, a variable $z_i$ with (fractional) value $a_i$ in the relaxation is selected for branching to two further nodes. In the right node, the constraint $z_i \geq a_i$ is imposed and in the left, $z_i < a_i$;  bisecting the search space and discarding fractional values in between the bounds. The solution of each LP relaxtion provides a lower bound on the MILP's optimal objective value. When an LP relaxation happens to admit an integral solution, an upper bound is obtained. The minimal upper bound found thus far is used, at each node, for pruning.

Finally, \emph{Constraint Programming} \citep{rossi2006handbook} is an additional effective paradigm adopted to solve industrial-sized MI(L)P and discrete combinatorial programs.

\section{Preliminaries: Deep Learning}





Supervised deep learning can be viewed as the task of
approximating a complex non-linear mapping from targeted data. Deep
Neural Networks (DNNs) are deep learning architectures composed of a
sequence of layers, each typically taking as inputs the results of the previous layer \citep{lecun2015deep}. Feed-forward neural networks are basic DNNs where the layers are fully connected and the function connecting the layer is given by
\(
    \bm{o} = \pi(\bm{W} \bm{x} + \bm{b}),
\)
where $\bm{x} \!\in\! \mathbb{R}^n$ and is the input vector, 
$\bm{o} \!\in\! \mathbb{R}^m$ the output vector, $\bm{W} \!\in\! 
\mathbb{R}^{m \times n}$ a matrix of weights, and $\bm{b} \!\in\! 
\mathbb{R}^m$ a bias vector. The function $\pi(\cdot)$ is
often non-linear (e.g., a rectified linear unit (ReLU)).

Supervised learning tasks consider datasets $\bm{\chi}\!=\!\{\bm{x}_i, \bm{y}_i\}_{i=1}^N$ consisting of $N$ data points with $\bm{x}_i \in \mathcal{X}$ being a feature vector and $\bm{y}_i \in \mathcal{Y}$ the associated targets. The goal is to learn a model ${\cal M}_\theta : \mathcal{X} \to \mathcal{Y}$, where $\theta$ is a vector of real-valued parameters, and whose quality is measured in terms of a nonnegative, and assumed differentiable, \emph{loss function} $\mathcal{L}: \mathcal{Y} \times \mathcal{Y} \to \mathbb{R}_+$.  The learning task minimizes the empirical risk function (ERM):
\begin{equation}
\label{eq:erm}
    \min_\theta J({\cal M}_\theta, \bm{\chi}) = \frac{1}{n} \sum_{i=1}^n 
    \mathcal{L}({\cal M}_\theta(\bm{x}_i), \bm{y}_i).
\end{equation}

Most of the techniques surveyed in this work use (variants of) DNNs whose training conforms to the objective above. Other notable classes of deep learning methods used to solve CO problems are Sequence Models and Graph Neural Networks (GNNs), which are reviewed below, and 
Reinforcement Learning (RL), which differs from supervised learning in not requiring labeled input/output pairs and concerns with learning a policy that maximizes some expected reward function.  
We refer the reader to 
\citep{sutton2018reinforcement} for an extensive overview of RL.

\paragraph{Sequence Models}
Recurrent neural networks (RNNs) have proven highly effective for tasks requiring pattern recognition in sequential data. A basic RNN uses a \emph{recurrent layer} to compute a sequence of output arrays $\bm{y}_1, \ldots, \bm{y}_N$ from a sequence of input arrays  $\bm{x}_1, \ldots, \bm{x}_N$. Each of several hidden units $h_t$ depends on its corresponding input array $\bm{x}_t$ along with the previous hidden unit $h_{t-1}$ in the following manner:
$$ 
\bm{h}_t = \pi( \bm{W}^x \bm{x}_t + \bm{W}^h \bm{h}_{t-1} +  \bm{b}) 
$$
where $\pi$ is a nonlinear activation function, $\bm{W}^x$,$\bm{W}^h$ are weight matrices, and $\bm{b}$ is an additive bias. $\bm{W}$ and $\bm{b}$ are learned parameters, applied at each unit of the recurrent layer. 

In their basic form, RNNs are known to suffer from vanishing gradient issues: the propagation of gradients through many hidden units by the chain rule can lead to loss of information as gradient magnitudes diminish. Modern RNNs use gated recurrent units (GRU) to form long-short term memory models (LSTM) which can better preserve relevant information across many hidden layers. 

\emph{Sequence-to-Sequence}  models build on recurrent neural networks with the aim of extending the framework to handle input and output sequences of variable size. This is accomplished by the introduction of an \emph{endoder-decoder} architecture in which an encoder LSTM maps an input sequence ${\bf x} = (\bm{x}_1,\ldots,\bm{x}_N)$ to a fixed-length context vector $\bm{c}$, by way of the hidden encoder states $(e_1,\ldots,e_N)$ (typically, $\bm{c} = e_N$). 
A subsequent decoder LSTM is used to map the context vector to a sequence of hidden decoder states $(d_1,\ldots,d_M)$, which are used to target a sequence of labeled vectors $\bm{y} = (\bm{y}_1,\ldots, \bm{y}_M)$. That is, some function $g$ is trained to approximate the likelihood of predicting $\bm{y}_t$ given the context vector $\bm{c}$, the current decoder state and all previous $\bm{y}_i$:
\begin{equation*}
	\Pr(\bm{y}_t | \{ \bm{y}_1, \ldots, \bm{y}_{t-1} \}, \bm{c})   = 
	g(\bm{y}_{t-1}, \bm{d}_t, \bm{c}),
\end{equation*}
where the overall training goal is to model the probability of predicting sequence $\bf y$. This joint probability is decomposed into a product of ordered conditionals: 
\begin{equation*}
	\Pr({\bf y}) = \Pi_{t=1}^M 
		\Pr(\bm{y}_t | \{ \bm{y}_1, \ldots, \bm{y}_{t-1} \}, \bm{c}).
\end{equation*} 

The primary downside to this approach is the reliance on a single fixed-length vector to encode the input sequence, resulting in limited expressivity, particularly when the input sequences observed at test time are of a different length than those used during training. 
\emph{Attention} mechanisms \citep{bahdanau2016neural} are used to address this shortcoming. They use the same encoder design, but allow the decoder to adaptively use any subset of the encoder's hidden units during decoding. 
In place of a single context vector, a \emph{sequence} of context vectors $\bm{c}_r$ (one per decoder unit) is computed as 
\begin{subequations}
\begin{flalign}
\label{eq:attention1}
    \bm{c}_r &= \textstyle \sum_{t = 1}^N \bm{a}_{rt} e_t  \quad \forall r \leq M\\
\label{eq:attention2}
   \bm{a}_{rt} &= \textstyle \frac{ \exp(\bm{u}_{rt}) }{  \sum_{k=1}^N \exp(\bm{u}_{rk})  }\\
\label{eq:attention3}
   \bm{u}_{rt} &= \bm{v}^T \pi(\bm{W}_1 e_t + \bm{W}_2 d_r), 
\end{flalign}
\end{subequations}
where $\bm{a}$ above is computed as the softmax of $\bm{u}$, the result of \emph{alignment} model (\ref{eq:attention3}) in which $\bm{v}$, $\bm{W}_1$, and $\bm{W}_2$ are learnable matrices and $\pi$ is a nonlinear activation function. The values $\bm{a}_{rt}$ constitute a sequence of \emph{attention} vectors which measure the alignment between decoder state $d_r$ and encoder state $e_t$. Each attention vector is used as a set of weights to form a linear combination of encoder states, used as a context vector to inform subsequent predictions.

\emph{Pointer networks} \citep{vinyals2017pointer}  are a special use case of attention mechanisms, used when the intended output of a network is a permutation of its input. Rather than utilizing the attention vector to assist output sequence prediction pointer networks treat the attention vector itself as the prediction. That is, 
\begin{align*}
	&\bm{u}_{rt} = \bm{v}^T \pi(\bm{W}_1 e_t + \bm{W}_2 d_r) 
	\quad \forall 1 \leq t \leq N, \\
	&\Pr(\bm{y}_t | \{ \bm{y}_1, \ldots, \bm{y}_{t-1} \}, \bm{x})  
	= \text{softmax}( \bm{u}_r ),
\end{align*}
where $\bm{x}$ is again the input sequence and $\bm{y}$ is a permutation of $\bm{x}$. The attention vector is viewed as a pointer to a particular encoder state, and can be used in subsequent predictions.

\paragraph{Graph Neural Networks}
Graph Neural Networks (GNNs) \citep{scarselli2008graph}, learn to represent each node $v$ of a graph with a \emph{state embedding} $x_v$, which depends on information from its neighboring nodes. The goal of these embeddings is to provide useful latent information about each node to subsequent components of a prediction model. Let $x_v, x_{co[v]}, h_{ne[v]}, x_{ne[v]}$ represent, respectively, the features of node $v$, the features of its edges, the state embeddings, and features of its neighbors. Then consider two functions: 
\begin{subequations}
\label{eq:gnn}
  \begin{flalign}
  \label{eq:gnn1}
     h_v &= f(x_v, x_{co[v]}, h_{ne[v]}, x_{ne[v]}) \\
	\label{eq:gnn2}
    o_v &= g( h_v, x_v)
  \end{flalign}
\end{subequations}
Equivalently to (\ref{eq:gnn1}) and (\ref{eq:gnn2}), 
\begin{subequations}
  \begin{flalign}
	\label{eq:gnn3}
    	H &= F(H,X)\\
	\label{eq:gnn4}
    	O &= G(H,X_N)
  \end{flalign}
\end{subequations}
where $X$  and $H$ represent the concatentation of all features and all state embeddings, and $X_N$ contains only the node-wise features. When $F$ is a contractive mapping, the equation has a fixed point solution $H^*$ which can be found by repeated application  of \eqref{eq:gnn3}. When $F$ and $G$ are interpreted as feedforward neural networks, their parameters can be trained to produce state embeddings that are optimized to assist a particular learning task.

Several variants and enhancements to the basic GNN have been proposed; \cite{zhou2018graph} provides a thorough survey of techniques and applications. 

\begin{figure*}[t]
  \centering
  \includegraphics[width=0.91\linewidth]{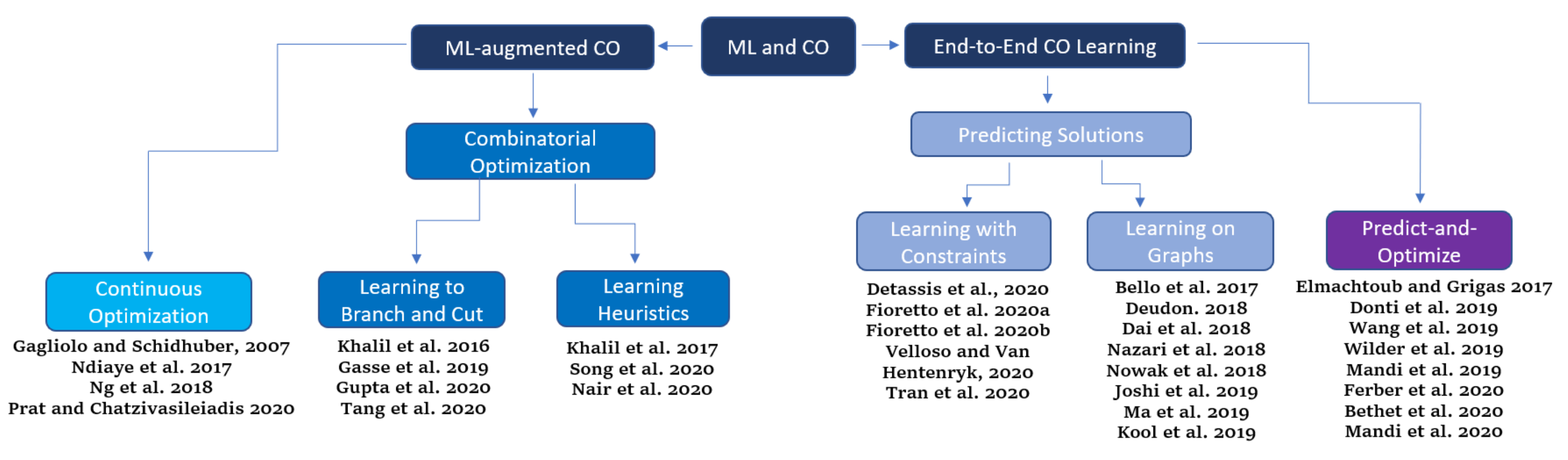}
  \caption{Machine Learning and Constrained Optimization.
  \label{fig:survey_overview}}
\end{figure*}

\section{Overview of ML and CO}
Current research areas in the synthesis of constrained optimization and machine learning can be categorized into two main directions: \emph{ML-augmented CO}, which focuses on using ML to aid the performance of CO problem solvers, and \emph{End-to-End CO learning (E2E-COL)}, which focuses on integrating combinatorial solvers or optimization methods into deep learning architectures.
A diagram illustrating the main research branches within this area is provided in Figure \ref{fig:survey_overview}. 

The area related with End-to-End CO learning is the focus of this survey and is concerned with the data-driven prediction of solutions to CO problems. This paper divides this area into:
{\bf (1)} approaches that develop ML architectures to predict fast, approximate solutions to predefined CO problems, 
further categorized into \emph{learning with constraints} and \emph{learning on graphs}, and
{\bf (2)} approaches that exploit CO solvers as  neural network layers for the purpose of structured logical inference,  referred to here as the \emph{Predict-and-Optimize} paradigm.

\section{ML-augmented CO}

ML-augmented CO involves the augmentation of already highly-optimized CO solvers with ML inference models.  
Techniques in this area draw on both supervised and RL frameworks to develop more efficient approaches to various aspects of CO solving for both continuous and discrete combinatorial problems. 

In the context of combinatorial optimization, these are broadly categorized into methods that learn to guide the search decisions in branch and bound solvers, and methods that guide the application of primal heuristics within branch and bound. The former include low-cost emulation of expensive branching rules in mixed integer programming \citep{khalilbranch,gasse2019exact,gupta2020hybrid}, prediction of optimal combinations of low-cost variable scoring rules to derive more powerful ones \citep{balcan2018learning}, and learning to cut \citep{tang2020reinforcement} in cutting plane methods within MILP solvers. The latter include prediction of the most effective nodes at which to apply primal heuristics \citep{khalillearnheur}, 
 and specification of primal heuristics such as the optimal choice of variable partitions in large neighborhood search \citep{SongLNS}. The reader is referred to the excellent surveys \citep{Lodi2017Jul,bengio2020machine} for a thorough account of techniques developed within ML-augmented combinatorial optimization.

Techniques in this area have also used ML to improve decisions in continuous CO problems and include learning restart strategies \citep{gagliolo2007learning}, learn rules to ignore some optimization variables leveraging the expected sparsity of the solutions and consequently leading to faster solvers, \citep{ndiaye2017gap}, and learning active constraints \citep{ng2018statistical,prat2020learning} to reduce the size of the problem to be fed into a CO solver
. 

\section{E2E-COL: Predicting CO Solutions}

A diverse body of work within the end-to-end CO learning literature has focused on developing ML architectures to predict fast, approximate   solutions to predefined CO problems \emph{end-to-end} without the use of CO solvers at the time of inference, by observing a set of solved instances or execution traces. These approaches contrasts with those that use ML to augment search-based CO solvers and configure their subroutines to direct the solver to find solutions efficiently. This survey categorizes the literature on \emph{predicting CO solutions} into {\bf (1)} methods that incorporate constraints into end-to-end learning, for the prediction of feasible or near-feasible solutions to both continuous and discrete constrained optimization problems, and {\bf (2)} methods that learn combinatorial solutions on graphs, with the goal of producing outputs as {combinatorial structures from variable-sized inputs}.
These two categories, referred to as \emph{learning with constraints} and \emph{learning CO solutions}, respectively, are reviewed next.

\subsection{Learning with Constraints}

The methods below consider datasets $\bm{\chi} = \{\bm{x}_i, \bm{y}_i\}_{i=1}^N$ whose inputs $\bm{x}_i$ describe some problem instance 
specification, such as matrix $\bm{A}$ and vector $\bm{b}$ describing linear constraints in MILPs, 
and the labels $\bm{y}_i$ describe complete solutions to problem ${\cal O}$ with input $\bm{x}_i$. 
Notably, each sample may specify a different problem instance (with 
different objective function coefficients and constraints).

An early approach to the use of ML for predicting CO problem solutions was presented by \cite{hopfield_tsp}, which used Hopfield Networks (\cite{hopfield_networks}) with modified energy functions to emulate the objective of a traveling salesman problem (TSP), and applied Lagrange multipliers to enforce feasibility to the problem's constraints. It was shown in \cite{wilson1988stability} however, that this approach suffers from several weakness, notably sensitivity to parameter initialization and hyperparameters.  As noted in \cite{bello2017neural} , similar approaches have largely fallen out of favor with the introduction of practically superior methods. 

Frameworks that exploit Lagrangian duality to guide the prediction of approximate solutions to satisfy the problem's constraints have found success in the context of continuous NLPs including energy optimization \citep{Fioretto:AAAI-20,velloso2020combining} as well as constrained prediction on tasks such as transprecision computing and fair classification \citep{fioretto2020lagrangian,tran:20_fairdp}.

Other end-to-end learning approaches have demonstrated success on the prediction  of solutions to constrained problems by injecting information about constraints from targeted feasible solutions. Recently, \cite{detassis2020teaching} presented an iterative process of using an external solver for discrete or continuous optimization to adjust targeted solutions to more closely match model predictions while maintaining feasibility, reducing the degree of constraint violation in the model predictions in subsequent iterations.

\subsection{Learning Solutions on Graphs}


By contrast to approaches learning solutions to unstructured CO problems, a variety of methods learn to solve CO cast on graphs. The development of deep learning architectures, such as sequence models and attention mechanisms, as well as GNNs, has provided a natural toolset for these tasks. 

The survey categorizes these modern approaches broadly based on whether they rely supervised learning or reinforcement learning. It observes that: {\bf (1)} Supervised learning concerns training models that take as input a representation of a problem instance and output a solution. Target solutions must be provided for this training task; this is often problematic as it entails solving, typically, NP-hard problems. {\bf (2)} Optimization problems have a native objective function, which in principle can be used in lieu of a loss function in supervised learning (possibly obviating the need for training labels) and can serve as a natural reward function in reinforcement learning. 


\subsubsection{Supervised Learning}

\cite{vinyals2017pointer} introduced the \emph{pointer network}, in which a sequence-to-sequence model uses an encoder-decoder architecture paired with an attention mechanism to produce permutations over inputs of variable size. Noting that certain CO problems provide a compelling use case for their architecture, the authors test it for solving the Traveling Salesman (TSP), Delaunay Triangulation, and Convex Hull problem variants. In each, the solution to a problem instance can be expressed as a single permutation. They develop a pointer network model to predict approximately optimal solutions by learning from previously solved instances in a supervised manner, and demonstrate some ability to generalize over variable-sized problem instances. 
For example, in the $2D$ Euclidean TSP, the pointer network's inputs are the $2D$ coordinates of each city that must be visited. A predicted permutation represents a tour over all cities, and each target label is a permutation representing the pre-computed minimum-length tour over its corresponding input coordinates. Any city is directly reachable from any other---by a straight line whose length is equal to the euclidean distance. In general, the approach introduced in this work applies only to problems whose solutions take the form of a single permutation and where all permutations are feasible (e.g., the problem does not feature constraints on which tours are allowed in the TSP).

Despite the subsequent trend toward RL-oriented frameworks, as discussed in the next section, \cite{nowak2018revised} studied a purely supervised learning method for the general Quadratic Assignment Problem (QAP). The proposed solution was based on the use of simple graph neural networks trained on representations of individual problem instances and their targeted solutions. The TSP is chosen as a test problem, along with two graph matching problems (all instances of the QAP).  Inferences from the model take the form of approximate permutation matrices, which are converted into feasible tours by a beam search. The authors note promising accuracy results on small instances, but the ability for trained models to generalize their performance to larger-sized instances was not shown.

\cite{joshi2019efficient} built on this concept by applying a Graph Convolutional Network model to the $2D$ Euclidean TSP. Aside from the enhanced deep learning architecture, most aspects of the approach remain similar. However, accurate results are reported on problem instances much larger than in \cite{nowak2018revised}.  The authors observe that compared to contemporary reinforcement learning approaches, their training method is more sample-efficient, but results in models that do not generalize nearly as well to problem instances larger than those used in training.

\subsubsection{Reinforcement Learning}

The pointer network architecture of \cite{vinyals2017pointer} was  adopted by \cite{bello2017neural}, who proposed to train an approximate TSP solver with reinforcement learning rather than supervised learning. The transition to RL was motivated partly by the difficulties associated with obtaining target solutions that are optimal, and the existence of nonunique optimal solutions to TSP instances. Rather than generating and targeting precomputed solutions, the authors present an \emph{actor-critic} RL framework using expected tour length $L(\pi | g)$ as the reward signal, where $g$ and $\pi$ represent a graph (problem instance) and a permutation (a tour over the graph):
\begin{equation}
\label{eq:belloreward}
    J(\theta | g) = \mathbb{E}_{\pi \sim p_{\theta}(\pi|g)}L(\pi | g).
\end{equation}

A policy, represented by the underlying pointer network with parameters $\theta$, is optimized using the well-known REINFORCE rule for computing policy gradients \citep{williams1992simple}:
\begin{equation}
\label{eq:bellograd}
    \nabla_{\theta} J(\theta | g) = \mathbb{E}_{\pi \sim p_{\theta}(\pi|g)} 
[ (L(\pi | g) - b(g) \nabla_{\theta} \log p_{\theta} \; (\pi | g)] 
\end{equation}
The total training objective is then the expectation of $J(\theta | g)$ over a distribution of graphs.
The policy gradient calculation requires a baseline function $b$ which estimates the expected reward. In this work, an auxiliary \emph{critic} network with its own set of parameters is trained to predict the expected tour length for any graph in supervised fashion, using empirically observed tours from the most recent policy as training labels. At the time of inference, two methods are available for producing tours from a trained model: {\bf (1)} A set of tours can be sampled from the stochastic policy by varying the softmax temperature parameter, from which the shortest is selected or {\bf (2)} an active search method which modifies the stochastic policy based on solutions sampled during inference.


\cite{nazari2018reinforcement} departed from the RNN encoder scheme used in \cite{bello2017neural} and \cite{vinyals2017pointer} arguing that when a problem's defining data do not adhere to a natural order (e.g. city coordinates in the TSP), it need not be modeled sequentially. This provides motivation to leave out the encoder RNN and consider combinatorial problems whose state changes over time, replacing the role of encoder hidden states with representations of the problem in each time frame. The authors consider a capacitated vehicle routing problem (VRP/CVRP) in which demands are associated to points in space and change over time; a vehicle with limited capacity must satisfy all the demands by delivering supply from a central depot while minimizing total tour length. This setting also diverges from those previously discussed in that solutions no longer take the form of permutations, since an optimal tour may now include several trips to the supply depot or a demand sink. 
The static (location coordinates) and dynamic (demand value) data defining the problem are concatenated into a single embedding vector $\bm{x}_t$ for each time frame $t$. In each decoder step, the attention layer computes a context vector $c_t$ based only on $\bm{x}_t$ and the previous decoder state $d_t$.  A final prediction at time $t$ indicates the next visited located based on $\bm{x}_t$ and $c_t$, and the embeddings $\bm{x}_t$ can then be updated based on the resulting state-changes. The training consists of an actor-critic method similar to that of \cite{bello2017neural}.

Several improvements of these approaches exists. Notably, 
\cite{kool2019attention} developed a general reinforcement learning framework based on a Graph Attention Network architecture \citep{velivckovic2017graph} and trained with the REINFORCE policy gradients, and present significant improvements in accuracy on 2D euclidean TSP  over \cite{vinyals2017pointer}, \cite{bello2017neural}, \cite{dai2018learning}, and \cite{nowak2018revised}.


Differently from previous approaches, \cite{ma2019combinatorial} 
focused on CO instances with hard constraints and consider TSP variants in which certain tours are considered infeasible. 
A mutil-level RL framework is proposed in which each layer of a hierarchy learns a different policy, from which actions can be sampled. Lower layers learn policies that result in feasible solutions by using reward functions that bias feasibility, and provide latent variables to higher layers which learn to optimize a given objective while sampling actions from lower layers. Layer - wise policy optimization uses the REINFORCE algorithm, training each layer in turn before fixing its weights and sampling actions to train the next layer. Each policy is represented by a graph neural network to produce context embeddings that capture inter-node relationships from node-wise inputs, and an attention-based decoder to predict distributions over candidate actions as in prior works. When the direct output $\bm{u}_i$ from layer $i$ is predicted at a particular decoding step, it is combined with the output of the lower layer $i-1$ to ensure mutual compatibility in the policy distribution
$$  \bm{p}_i = \textrm{softmax}( \bm{u}_i + \alpha \bm{u}_{i-1} )  $$ where $\alpha$ is a trainable parameter.

Finally, \cite{dai2018learning} adopted a different RL approach based on a greedy heuristic framework which determines approximate solutions by the sequential selection of graph nodes. The selection policy is learned using a combination of Q-learning and fitted Q-iteration, by optimizing the optimization objective directly. The graph embedding network \texttt{structure2vec} is used to represent the policy of the learned greedy algorithm. It is used to produce feature embeddings for each node of the graph, which are updated between actions and capture relationships between neighboring nodes, given the current state of a partial solution. It is noted that this framework excels in combinatorial problems that have a true graphical structure, as opposed to most previous previously studied applications (primarily the TSP and VRP variants) in the underlying 'graph' is typically fully connected. The Minimum Vertex Cover, Max Cut, and TSP are used as test test problems, and a strong ability to generalize over various problem sizes is advertised in comparison to prior approaches.

\begin{figure}[!t]
  \centering
  \includegraphics[width=1.0\linewidth]{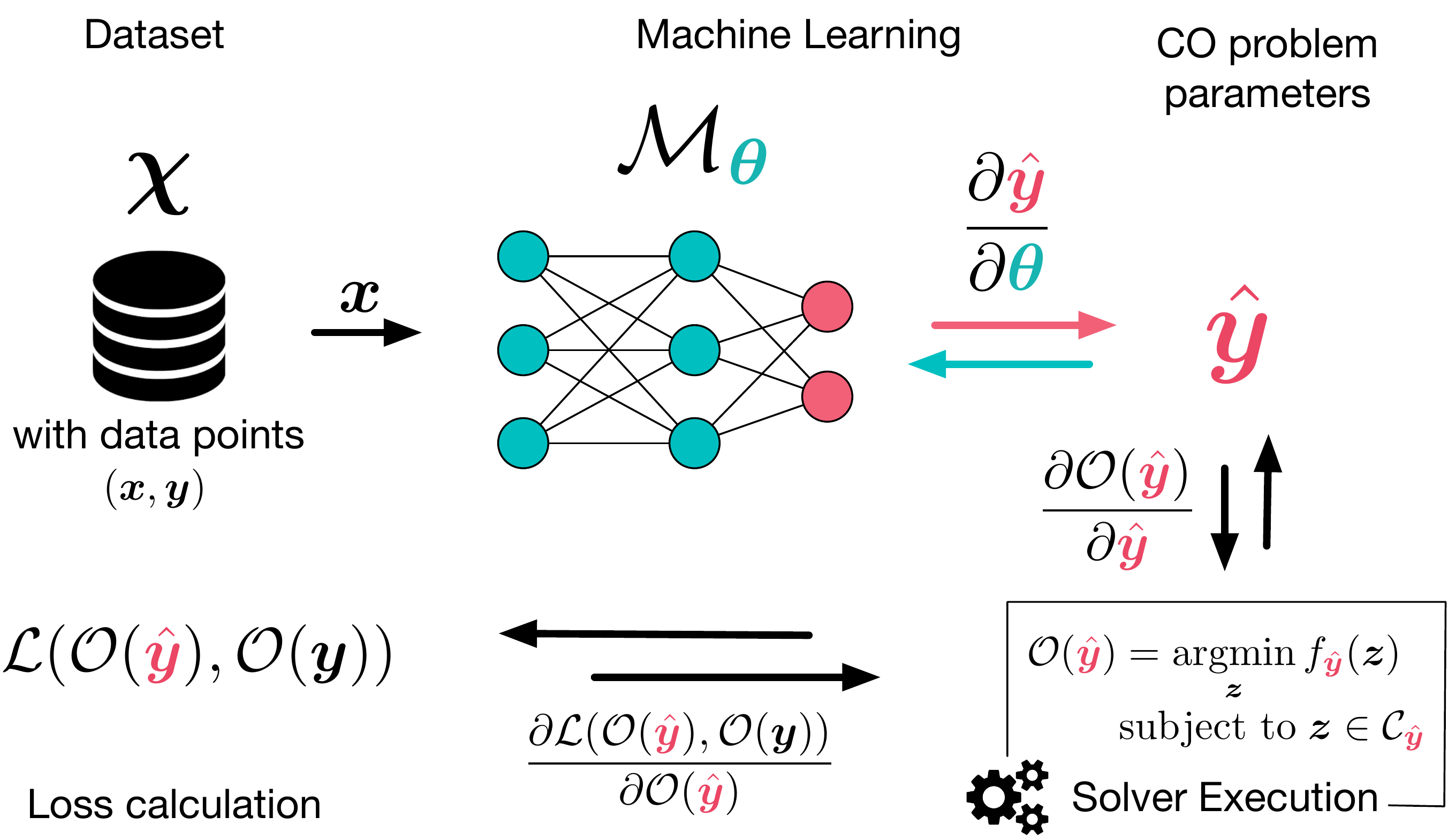}
  \caption{Predict-and-optimize framework; gradients of a solver output (solution) must be computed with respect to its input (problem parameters) in order to maximize empirical model performance.
    \label{fig:pred_opt_flowchart}}
\end{figure}


\section{E2E-COL: Predict-and-Optimize}

A burgeoning topic in the intersection of ML and CO is the fusion of prediction (ML) and decision (CO) models, in which  decision models are represented by partially defined optimization problems, whose specification is completed by parameters that are predicted from data. The resulting composite models employ constrained optimization as a neural network layer and are trained \emph{end-to-end}, based on the optimality of their decisions. This setting is altogether different in motivation to those previously discussed, in which the goal was to solve predefined CO instances with increased efficiency. The goal here is the synthesis of predictive and prescriptive techniques to create ML systems that learn to make decisions based on empirical data.

The following constrained optimization problem is posed, in which the objective function $f_{\bm{y}}$ and feasible region ${\cal C}_{\bm{y}}$ depend on a parameter vector $\bm{y}$:
\begin{equation}
\label{eq:opt_theta}
\mathcal{O}(\bm{y}) = \argmin_{\bm{z}} f_{\bm{y}}(\bm{z}) \;\; 
  \text{subject to} \;\;
  \bm{z} \in \mathcal{C}_{\bm{y}}.
\end{equation}
The goal here is to use supervised learning to predict $\hat{\bm{y}}$ the unspecified parameters from empirical data. The learning task is performed so that the optimal solution ${\cal O}(\hat{\bm{y}})$ best matches a targeted optimal solution ${\cal O}(\bm{y})$, relative to some appropriately chosen loss function. The empirical data in this setting is defined abstractly as belonging to a dataset $\chi$, and can represent any empirical observations correlating with targeted solutions to \eqref{eq:opt_theta} for some $\bm{y}$. See Figure \ref{fig:pred_opt_flowchart} for an illustration.

This framework aims to improve on simpler two-stage approaches, in which a conventional loss function (e.g. MSE) is used to target labeled parameter vectors $\bm{y}$ that are provided in advance, before solving the associated CO problem to obtain a decision. Such approaches are suboptimal in the sense that predictions of $\bm{y}$ do not take into account the accuracy of the resulting solution ${\cal O}(\bm{y})$ during training.

We note that there are two ways to view the predictions that result from these integrated models. If $\hat{\bm{y}}$ is viewed as the prediction, then the calculation of ${\cal O}(\hat{\bm{y}})$ is absorbed into the loss function $\mathcal{L}(\hat{\bm{y}},\bm{y})$, which targets the provided parameter vectors. Otherwise, the loss function $\mathcal{L}( {\cal O}(\hat{\bm{y}}) , {\cal O}(\bm{y}))$ is considered separately from the decision model and aims to match computed optimal solutions to targeted ones. 
One advantage sought in either case is the opportunity to minimize during training the ultimate error in the computed optimal objective values $f_{\hat{\bm{y}}}({\cal O}(\hat{\bm{y}}))$, relative to those of the target data. This notion of training loss is known as \emph{regret:
$$
\textsl{regret}(\hat{\bm{y}},\bm{y}) = 
  f_{\hat{\bm{y}}}({\cal O}(\hat{\bm{y}}))-f_{\bm{y}}({\cal O}(\bm{y})).
$$
Otherwise the optimal solution ${\cal O}(\bm{y})$ is targeted and one can use
$\textsl{regret}({\cal O}(\hat{\bm{y}}),{\cal O}(\bm{y}))$, regardless of whether the corresponding $\bm{y}$ is available.} 
Depending on the techniques used, it may be possible to minimize the regret without access to ground-truth solutions, as in \cite{wilder2018melding}, since the targeted solutions ${\cal O}(\bm{y})$ contribute only a constant term to the overall loss.
It is worth mentioning that because the optimization problem in \eqref{eq:opt_theta} is viewed as a function, the existence of nonunique optimal solutions is typically not considered. The implication then is that ${\cal O}(\bm{y})$ is directly determined by $\bm{y}$.

Training these end-to-end models involves the introduction of external CO solvers into the training loop of a ML model, often a DNN. Note that combinatorial problems with discrete state spaces do not offer useful gradients; viewed as a function, the \emph{argmin} of a discrete problem is piecewise constant.     
\emph{The challenge of forming useful approximations to  
$\frac{\partial \mathcal{L}}{\partial y } $ is
 central in this context and must be addressed in order to perform backpropagation}. 
 It may be approximated directly, but a more prevalent strategy is to model $\frac{\partial {\cal O}(y)}{\partial y } $ and $ \frac{\partial \mathcal{L}}{\partial {\cal O}}$ separately, in which case the challenge is to compute the former term by \emph{differentiation through argmin}.  Figure \ref{fig:pred_opt_flowchart} shows the role of this gradient calculation in context.

\subsection{Quadratic Programming}
One catalyst for the development of this topic was the introduction of \emph{differentiable optimization layers}, beginning with \cite{amos2019optnet} which introduced a GPU-ready QP solver that offers exact gradients for backpropagation by differentiating the KKT optimality conditions of a quadratic program at the time of solving, and utilizing information from the forward pass to solve a linear system for incoming gradients, once the outgoing gradients are known. Subsequently,  \cite{donti2019taskbased} proposed a \emph{predict-and-optimize} model in which QPs with stochastic constraints were integrated in-the-loop to provide accurate solutions to inventory and power generator scheduling problems specified by empirical data.

\subsection{Linear Programming}

Concurrent with \cite{donti2019taskbased}, an alternative methodology for end-to-end learning with decision models, called \emph{smart predict-and-optimize} (SPO), was introduced by \cite{elmachtoub2020smart}, which focused on prediction with optimization of constrained problems with linear objectives, in which the cost vector is predicted from data and the feasible region $\mathcal{C}$ is invariant to the parameter $\bm{y}$:
\begin{equation}
\label{eq:opt_spo}
\mathcal{O}(\bm{y}) = \argmin_{\bm{z}} \bm{y}^T\bm{z} \;\; 
  \text{subject to} \;\;
  \bm{z} \in \mathcal{C}.
\end{equation}
The target data in this work are the true cost vectors $\bm{y}$, and an inexact subgradient calculation is used for the backpropagation of regret loss 
$\mathcal{L}(\hat{\bm{y}}, \bm{y}) = \hat{y}^T ( {\cal O}(\hat{\bm{y}}) - {\cal O}(\bm{y})  )$  on the decision task, by first defining a convex surrogate upper bound on regret called the \emph{SPO+loss}, for which it is shown that ${\cal O}(\bm{y})-{\cal O}(2\hat{\bm{y}} - \bm{y})$ is a useful subgradient. 
Since this work is limited to the development of surrogate loss functions on regret from the optimization task, it does not apply to learning tasks in which the full solution to an optimization problem is targeted. The paper includes a discussion justifying the method's use on problems with discrete constraints in $\mathcal{C}$, as in combinatorial optimization, but experimental results are not provided on that topic. It is, however, demonstrated that the approach succeeds in a case where $\mathcal{C}$ is convex but nonlinear.  

\cite{wilder2018melding} introduced an alternative framework to predict-and-optimize linear programming problems, based on exact differentiation of a smoothed surrogate model. While LPs are special cases of QPs, the gradient calculation of \cite{amos2019optnet} does not directly apply due to singularity of the KKT conditions when the objective function is purely linear. This is addressed by introducing a small quadratic regularization term to the LP objective $f_{\bm{y}}(\bm{z}) = \bm{y}^T \bm{z}$ so that the problem in \eqref{eq:opt_theta} becomes
\begin{equation}
\label{eq:opt_theta_LP}
\mathcal{O}(\bm{y}) = \argmin_{\bm{z}} \bm{y}^T \bm{z} + \epsilon\|\bm{z}\|^2 \;\;
  \text{subject to} \;\;
  \bm{A}\bm{z} \leq \bm{b}.
\end{equation}
The resulting problems approximate the desired LP, but have unique solutions that vary smoothly as a function of their parameters, allowing for accurate backpropagation of the result. 
The integrated model is trained to minimize the expected optimal objective value across all training samples, which is equivalent to minimizing the regret loss but without the need for a target dataset.
 This work demonstrated success on problems such as the knapsack (using LP relaxation) and bipartite matching problems where a cost vector is predicted from empirical data (e.g., historical cost data for knapsack items), and is shown to outperform two-stage models which lack integration of the LP problem. We note that although the differentiable QP solving framework of \cite{amos2019optnet} is capable of handling differentiation with respect to any objective or constraint coefficient, this work only report results on tasks in which the \emph{cost} vector is parameterized within the learning architecture, and constraints are held constant across each sample. 
\emph{This limitation is common to all of the works described below, as well.} 

Next, \citet{mandi2020interior} introduced an altogether different approach to obtaining approximate gradients for the {\em argmin} of a linear program. An interior point method is used to compute the solution of a homogeneous self-dual embedding with early stopping, and the method's log-barrier term is recovered and used to solve for gradients in the backward pass. Equivalently, this can be viewed as the introduction of a log-barrier regularization term, by analogy to the QP-based approach of \cite{wilder2018melding}:
\begin{equation*}
\label{eq:opt_theta_QP}
\mathcal{O}(\bm{y}) = \argmin_{\bm{z}} \bm{y}^T\bm{z} + 
  \lambda\left(\sum_i \ln (z_i) \right) \;\;
  \text{subject to} \;\;
  \bm{A}\bm{z} \leq \bm{b}.
\end{equation*}
Further, the method's performance on end-to-end learning tasks is evaluated against the QP approach of \cite{wilder2018melding} on LP-based predict-and-optimize tasks, citing stronger accuracy results on energy scheduling and knapsack problems with costs predicted from data.

\cite{berthet2020learning} detailed an approach based on stochastic perturbation to differentiate the output of linear programs with respect to their cost vectors. The output space of the LP problem is smoothed by adding low-amplitude random noise to the cost vector and computing the expectation of the resulting solution in each forward pass. This can be done in Monte Carlo fashion and in parallel across the noise samples. The gradient calculation views the solver as a black box in this approach, and does not require the explicit solving of LP for operations that can be mathematically formulated as LP, but are simple to perform (e.g., sorting and ranking). Results include a replication of the shortest path experiments presented in \citep{vlastelica2020differentiation}, in which a model integrated with convolutional neural networks is used to approximate the shortest path through stages in a computer game, solely from images.

\subsection{Combinatorial Optimization}

\cite{ferbermipaal} extended the approach of \cite{wilder2018melding} to integrate MILP within the end-to-end training loop, with the aim of utilizing more expressive NP-Hard combinatorial problems with parameters predicted from data. This is done by reducing the MILP with integer constraints to a LP by a method of cutting planes. In the ideal case, the LP that results from the addition of cutting planes has the same optimal solution as its mixed-integer parent. Exact gradients can then be computed for its regularized QP approximation as in \cite{wilder2018melding}. Although the LP approximation to MILP improves with solving time, practical concerns arise when the MILP problem cannot be solved to completion. Each instance of the NP-Hard problem must be solved in each forward pass of the training loop, which poses clear obstructions in terms of runtime. 
A disadvantage of the approach is that cutting-plane methods are generally considered to be inferior in efficiency to staple methods like branch and bound.
Improved results were obtained on portfolio optimization and diverse bipartite matching problems,  when compared to  LP-relaxation models following the approach of \cite{wilder2018melding}.

\cite{mandi2019smart} investigates the application of the same SPO approach to NP-Hard combinatorial problems. Primary among the challenges faced in this context is, as in \citep{ferbermipaal}, the computational cost of solving hard problems within every iteration of the training. The authors found that continuous relaxations of common MILP problems (e.g. knapsack) often offer subgradients of comaparable quality to the full mixed-integer problem with respect to the SPO loss, so that training end-to-end systems with hard CO problems can be simplified in such cases by replacing the full problem with an efficiently solvable relaxation, an approach termed \emph{SPO-relax}. The authors put continuous relaxations into the broader category of ``weaker oracles'' for the CO solver, which also includes approximation algorithms (e.g. greedy approximation for knapsack). 
The main results showed that SPO-relax achieves accuracy competitive with the full SPO approach but with shorter training times on a handful of discrete problems. The SPO-relax approach was compared also against the formulation of \cite{wilder2018melding} on equivalent relaxations, but no clear winner was determined.

\cite{vlastelica2020differentiation} introduced a new idea for approximating gradients over discrete optimization problems for end-to-end training, which relies on viewing a discrete optimization problem as a function of its defining parameters (in this context coming from previous layers), whose range is piecewise constant. The only requirement is that the objective be linear. The gradient calculation combines the outputs of two calls to an optimization solver; one in the forward pass on initial parameters $\bm{y}$, and one in the backward pass on perturbed parameters $\bar{\bm{y}}$. The results are used to construct a piecewise \emph{linear} function which approximates the original solver's output space, but has readily available gradients. Because the gradient calculation is agnostic to the implementation of the solver, it is termed "black-box differentiation". As such, input parameters to the solver do not correspond explicitly to coefficients in the underlying optimization problem. Results on end-to-end learning for the shortest path problem, TSP and min-cost perfect matching are shown. In each case, the discrete problem's specification is implicitly defined in terms of images, which are used to predict parameters of the appropriate discrete problem through deep networks. The optimal solutions coming from blackbox solvers are expressed as binary indicator matrices in each case and matched to targeted  optimal solutions using a Hamming distance loss function.

Finally,  \cite{wang2019satnet} presented a differentiable solver for the MAXSAT problem, another problem form capable of representing NP-Hard combinatorial problems. Approximate gradients are formed by first approximating the MAXSAT problem as a related semidefinite program (SPD), then differentiating its solution analytically during a specialized coordinate descent method \citep{wang2018mixing} which solves the SDP. The successful integration of MAXSAT into deep learning is demonstrated with a model trained to solve sudoku puzzles represented only by handwritten images.

\section{Challenges and Research Directions}

The current state of the art in integrating combinatorial optimization with end-to-end machine learning 
shows promise on challenging tasks for which there was previously no viable approach. Further, it has been demonstrated that a variety of non-overlapping approaches  can be effective.  Despite these encouraging results, a number of challenges remain that must be addressed to allow an integration that lives up to its full potential.
{\bf (1)} Despite the variety of approaches, the success of the predict-and-optimize paradigm has been demonstrated on a relatively limited set of optimization problems and a majority of reported experimental results focus on linear programming formulations. 
Challenges posed by the parametrization of constraints stand in the way of broader applications, but have not been yet been addressed. 
{\bf (2)} Issues associated with the runtime of combinatorial solvers in-the-loop still make some potential applications impractical. 
{\bf (3)} Additionally, despite being possible in theory, the role of the CO model in-the-loop has not been generalized successfully beyond being applied as the final layer of a deep model. The use of additional layers beyond the solver, or even compositions of CO solving layers, could potentially lead to new applications if the practical challenges were to be overcome.  
{\bf (4)} In predicting solutions to CO problems, the current methods cannot reliably guarantee the problem constraints to be satisfied. This critical shortcoming may be addressed by integrating ML approaches with methods from the robust optimization literature or by developing ad-hoc layers to project the predictions onto the feasible space. {\bf (5)} While it has been observed in limited contexts \cite{demirovic2019investigation} that predict-and-optimize frameworks based on optimization layers are competitive only when the underlying constrained problem is convex, this area still lacks theoretical results providing guarantees on their viability or performance.
Finally, {\bf (6)} the need for uniform benchmark experiments and systematic comparisons between each predict-and-optimization framework is apparent. \cite{demirovic2019investigation} provided a study comparing the approaches of \cite{wilder2018melding} and \cite{elmachtoub2020smart}, along with problem-specific approaches, on knapsack problems but did not conclude strongly as to which method should be preferred. Further, \cite{mandi2019smart} reports that for knapsack problems, SPO performs comparably on the knapsack problem whether the LP relaxation or the full problem is used, but does not show that this effect generalize to other NP-Hard problems. This signals a need for comprehensive studies that test performance on a variety of hard CO problems.

Although the approaches surveyed are still in an early stage of their development, and have been adopted mainly for academic purposes, we strongly believe that the integration between combinatorial optimization and machine learning is a promising direction for the development of new, transformative, tools in combinatorial optimization and learning.

\section*{Acknowledgments}

This research is partially supported by NSF grant 2007164. Its views and conclusions are those of the authors only and should not be interpreted as representing the official policies, either expressed or implied, of the sponsoring organizations, agencies, or the U.S.~government.

\bibliographystyle{named}
\bibliography{ijcai21}

\end{document}